\documentclass{article}
\usepackage[T1]{fontenc} % Keep the official Times fonts under Tectonic/XeTeX.
\usepackage{iclr2027_conference,times}
\usepackage{amsmath,amssymb}
\usepackage{booktabs}
\usepackage{graphicx}
\usepackage{placeins}
\usepackage{hyperref}
\usepackage{url}

\title{Growing and Transferring Pathways for LLM Reinforcement Learning}
\author{
Haolin Tian$^{1*}$ \quad Yikang Li$^{1*}$ \quad Tonghan Wang$^{1\dagger}$\\
\normalfont $^{1}$Tsinghua University
}

\iclrfinalcopy

\begin{document}
\maketitle
\fancyhead{}
\renewcommand{\headrulewidth}{0.4pt}
\begingroup
\renewcommand{\thefootnote}{\fnsymbol{footnote}}
\footnotetext[1]{These authors contributed equally.}
\footnotetext[2]{Corresponding author. Correspondence: \texttt{thw@mail.tsinghua.edu.cn}.}
\endgroup
\begin{abstract}
Reinforcement learning (RL) can substantially improve the reasoning capabilities of large language models, but an effective training run requires considerable resources. Reusing the computation acquired through RL across models would extend the value of that training. We propose GrowRL, an RL method that organizes learned updates into compact computational pathways through structural growth. GrowRL freezes the pretrained backbone and adds an independent causal attention branch and a feed-forward network branch at each Transformer layer. Zero-initialized output projections preserve the pretrained policy at initialization, and RL updates only the added pathways. The learned pathways remain explicit after training, allowing transfer to structurally compatible target models and subsequent adaptation with the target backbone frozen. The architecture also supports sequential transfer of additional pathways. By keeping the learned attention and feed-forward computation separate from the pretrained weights, GrowRL provides an explicit structure for preserving and reusing the outcome of an RL run. Experiments across five backbones show competitive mathematical reasoning and out-of-distribution performance with a small number of trainable parameters. Learned pathways provide immediate gains across all three transfer directions, with performance after only 10 target-side updates already close to that after 100 updates in every case. These results support computational pathways as parameter-efficient and reusable units of reinforcement learning adaptation.
\end{abstract}

% !TeX root = ../main.tex
\section{Introduction}
\label{sec:introduction}

Reinforcement learning with verifiable rewards (RLVR) has become an important approach for improving the reasoning performance of large language models \citep{growrl_deepseekr1,growrl_drgrpo}. An effective RL run can produce substantial gains, but requires considerable training resources. Its value would extend beyond the model being trained if the computation acquired during that run could also benefit other models. Full-parameter reinforcement learning embeds updates throughout the trained backbone, while parameter-efficient methods preserve them in a smaller set of parameters \citep{growrl_adapters,growrl_lora,growrl_dora}. A separately stored update, however, is not necessarily useful in another model: the computation it performs must remain effective on representations produced by the receiving backbone. Reusable RL updates therefore face two linked requirements. They must improve the original model within a limited trainable budget and retain value after transfer to a compatible model.

We propose \textbf{GrowRL}, a reinforcement learning method that organizes learned updates into compact computational pathways. GrowRL keeps the pretrained backbone frozen and adds a trainable pathway to each Transformer layer \citep{growrl_transformer}. Each pathway contains an independent causal attention branch and a compact feed-forward network (FFN) branch, whose outputs are added to those of the corresponding original sublayers. The output projections of the new branches are zero-initialized, preserving the original model's behavior at the start of training. Reinforcement learning optimizes only the added pathways, leaving the learned components separate from the backbone parameters. After training, the pathways can be inserted into corresponding layers of a structurally compatible target model and further adapted there. Their attention and feed-forward structure provides a consistent interface for transferring learned computation between structurally compatible backbones.

We evaluate GrowRL on multiple backbones and reasoning tasks, examining both learning on the original host and reuse in another model. On their original hosts, the pathways yield substantial reasoning improvements while training only a small number of added parameters, achieving performance competitive with full-parameter reinforcement learning across multiple settings. We then transfer trained pathways across three pairs of structurally compatible models and evaluate the receiving models before any target-side updates. In all three pairs, the transferred pathway improves MATH500 Pass@1 over the unmodified receiving model. Further reinforcement learning on the transferred pathway produces additional gains on Qwen2.5-1.5B and Qwen2.5-7B, while the pathway retains its initial advantage on Llama-3.2-3B. These results show that a pathway learned through one RL run can benefit its original host and remain useful after insertion into another compatible model.

GrowRL makes the computation acquired through reinforcement learning explicit in the model structure. The learned pathways can be retained independently of the original backbone and used as starting points for adaptation in a receiving model. The same structural design also permits further pathways to be added to an already adapted host. Together, these properties offer a way to preserve and reuse the outcome of an RL run across successive stages of model adaptation.

Our main contributions are as follows.
\begin{enumerate}
    \item We propose \textbf{GrowRL}, which performs reinforcement learning through newly introduced attention and feed-forward pathways while keeping the pretrained backbone frozen. The learned updates remain identifiable components that can be inserted into structurally compatible models.
    \item We show that compact GrowRL pathways can substantially improve reasoning performance on their original hosts and achieve performance competitive with full-parameter reinforcement learning across multiple settings.
    \item We evaluate immediate cross-model transfer and subsequent target-side adaptation. Across three source--target pairs, transferred pathways improve MATH500 performance without target-side training and remain useful during further adaptation.
\end{enumerate}

% !TeX root = ../main.tex
\section{Related Work}

\subsection{Reinforcement Learning and Parameter-Efficient Adaptation}
Reinforcement learning with verifiable rewards has become an important approach for improving the reasoning capabilities of large language models and has achieved substantial performance gains on mathematical and other reasoning tasks \citep{growrl_deepseekr1,growrl_drgrpo}. Meanwhile, parameter-efficient adaptation has developed into a rich line of research, including adapters \citep{growrl_adapters}, Low-Rank Adaptation (LoRA) \citep{growrl_lora}, Weight-Decomposed Low-Rank Adaptation (DoRA) \citep{growrl_dora}, and other adaptation structures \citep{growrl_pa}. These methods demonstrate that model updates can be concentrated within a relatively small parameter space. Related studies, including Tina, have also shown that a limited number of trainable parameters can effectively carry reinforcement learning updates and yield substantial improvements in reasoning performance \citep{growrl_tina}. How trainable parameters are organized is therefore an important design dimension in reinforcement learning for model adaptation.

\subsection{Modular Learning and Computational Structure Expansion}
Modular learning focuses on preserving acquired capabilities in computational units with explicit structural boundaries. Adapters and their composition methods demonstrate the training and composition of independent modules \citep{growrl_adapters,growrl_adapterfusion}. Progressive Networks acquire successive task capabilities, including in RL settings, by introducing new network structures \citep{growrl_progressive}, while LLaMA-Pro increases computational capacity in existing language models by expanding Transformer blocks \citep{growrl_llamapro}. Tiny-Attention Adapter introduces compact attention modules, providing another approach to adapting contextual computation \citep{growrl_tinyattention}.

These studies establish that models can be adapted or expanded through newly introduced computational structures, with applications to task adaptation, module composition, and continual learning. GrowRL focuses on preserving and reusing the computation acquired through an RL run. It organizes reinforcement learning updates into attention and feed-forward computational pathways at each layer. These pathways remain structurally distinct from the frozen backbone after training, allowing the learned computation to be transferred to compatible models, further adapted, and incorporated into an already adapted host through sequential transfer.

% !TeX root = ../main.tex
\section{GrowRL}
\label{sec:method}

\begin{figure}[t]
\centering
\includegraphics[width=\linewidth]{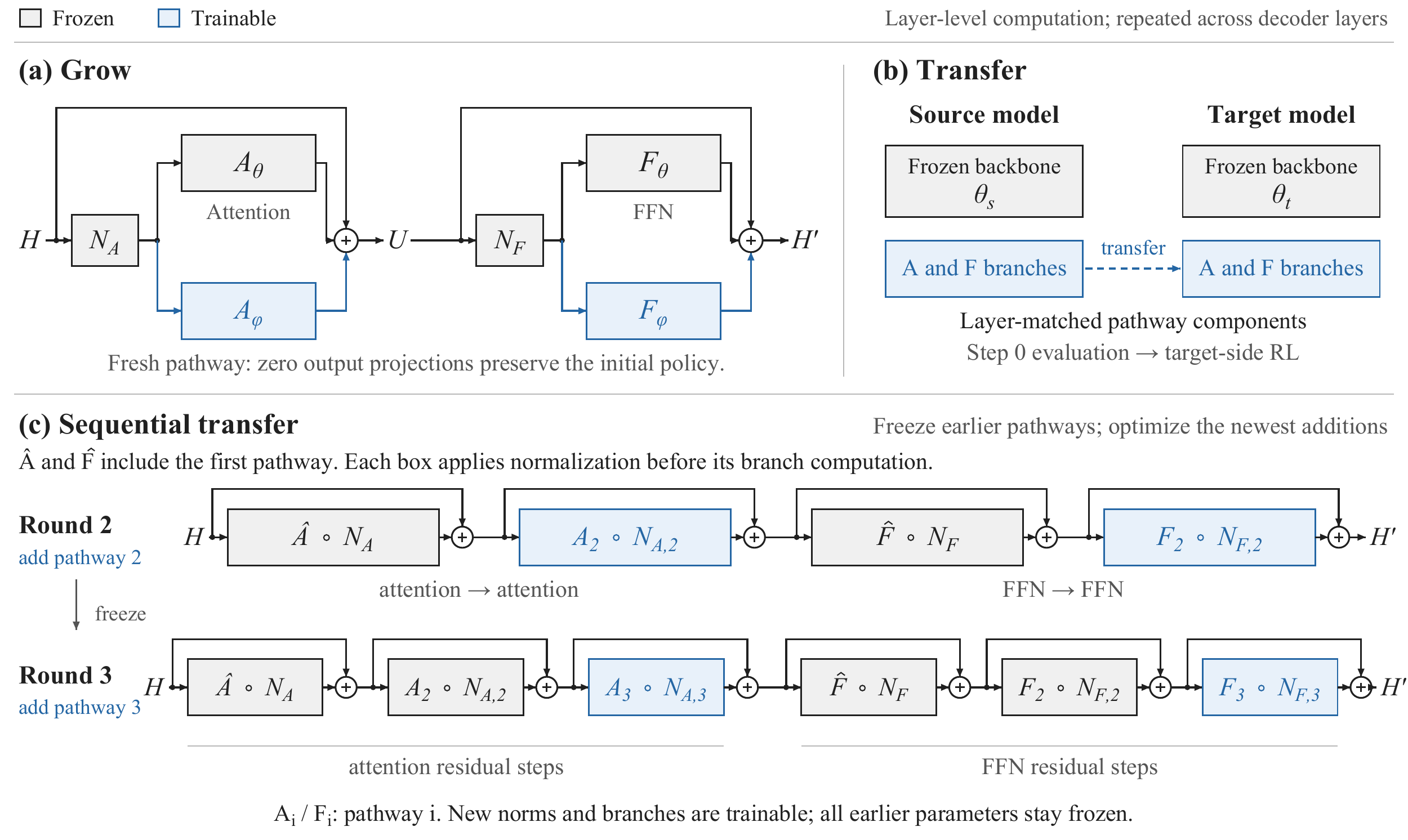}
\caption{\textbf{GrowRL overview.} (a) Train a compact pathway on a frozen backbone. (b) Transfer the learned pathway to a compatible target model. (c) Insert further pathways through sequential transfer.}
\label{fig:method_overview}
\end{figure}

GrowRL learns a computational pathway on a frozen source backbone and transfers the learned pathway to a structurally compatible target backbone. Let $\theta$ denote backbone parameters and $\phi$ pathway parameters. The pathway adds an attention branch and an FFN branch to each Transformer layer. Figure~\ref{fig:method_overview} illustrates source-model training, single transfer, and sequential transfer; $A_\theta,F_\theta$ denote backbone sublayers, and $A_i,F_i$ denote the branches of pathway $i$.

\suppressfloats[t]

\subsection{Structural growth and source-model training}
\label{sec:pathway}

Consider a pre-normalized decoder with $L$ layers and hidden dimension $d$. At one layer, let $H\in\mathbb{R}^{T\times d}$ contain the residual states of $T$ tokens. We suppress the layer index and denote the backbone attention and feed-forward sublayers by $A_\theta$ and $F_\theta$, with normalizations $N_A$ and $N_F$. GrowRL adds branches $A_\phi$ and $F_\phi$ that share the normalized inputs of the corresponding backbone sublayers:
\begin{equation}
\begin{aligned}
X &= N_A(H), & U &= H + A_\theta(X) + A_\phi(X), \\
Z &= N_F(U), & H' &= U + F_\theta(Z) + F_\phi(Z).
\end{aligned}
\label{eq:grow_layer}
\end{equation}
The added attention branch has its own query, key, value, and output projections. For all backbones considered here, it uses rotary positional embeddings (RoPE) \citep{growrl_rope}. With one head of dimension $d_h$, it computes
\begin{equation}
\begin{gathered}
Q = \operatorname{RoPE}(XW_Q), \qquad
K = \operatorname{RoPE}(XW_K), \qquad V = XW_V, \\
A_\phi(X) = \operatorname{softmax}\!\left(\frac{QK^\top}{\sqrt{d_h}} + M\right)VW_O.
\end{gathered}
\label{eq:grow_attention}
\end{equation}
Here $W_Q,W_K,W_V\in\mathbb{R}^{d\times d_h}$, $W_O\in\mathbb{R}^{d_h\times d}$, and $M$ masks future and padding positions. RoPE follows the backbone's positional convention, while the branch computes its own attention distribution. The FFN branch is a narrow SwiGLU transformation \citep{growrl_swiglu}:
\begin{equation}
F_\phi(Z) = \big[\operatorname{SiLU}(ZW_g)\odot(ZW_u)\big]W_d.
\label{eq:grow_ffn}
\end{equation}
Here $\operatorname{SiLU}(x)=x\,\sigma(x)$, where $\sigma$ is the sigmoid function, and $W_g,W_u\in\mathbb{R}^{d\times r}$ and $W_d\in\mathbb{R}^{r\times d}$. We use $d_h=r=128$ in the evaluated models. Both branches return to the $d$-dimensional residual stream, so subsequent backbone layers process the representations modified by the pathway.

\paragraph{Initialization and parameter budget.}
We initialize $W_Q,W_K,W_V,W_g,W_u$ from $\mathcal{N}(0,0.02^2)$ and set $W_O=W_d=0$. The added branches therefore output zero initially, preserving the backbone's policy in exact arithmetic. Initially, only the output projections can receive nonzero loss gradients; gradients reach the internal projections after the output projections change (Appendix~\ref{app:initialization}). We train the added pathways while keeping the backbone projections, normalizations, embeddings, and language-model head fixed. The branch matrices contain $L(4dd_h+3dr)$ parameters, giving approximately 38.5M trainable parameters for the 1.5B models.

\paragraph{Source-model reinforcement learning.}
\label{sec:rl_objective}
We train the pathway using Dr.~GRPO \citep{growrl_drgrpo}, sampling eight responses per prompt and assigning binary rewards from final-answer correctness. The backbone remains frozen, but its hidden states can change because later backbone layers process the pathway's outputs. GrowRL thus changes the trainable computational structure, while using the same reasoning objective and answer-verification reward as the comparison methods. The complete objective and implementation details appear in Appendices~\ref{app:rl_objective} and~\ref{app:implementation}.

\subsection{Transfer and Sequential Transfer}
\label{sec:transferring}

\paragraph{Transfer to a target backbone.}
After source-model training, we insert the learned pathway into the corresponding layers of a structurally compatible target backbone. Compatibility requires matching insertion sites and tensor dimensions for the transferred projections; no learned cross-model interface is used. The target retains its own embeddings, normalizations, attention and FFN parameters, and language-model head. Only pathway components are transferred. Transfer and target-side evaluation settings are given in Appendix~\ref{app:transfer}.

A trained pathway can change the target policy immediately upon insertion, without target-side optimization. We evaluate this zero-update model to measure immediate transfer. We then optionally optimize the inserted pathway through further RL while keeping the target backbone frozen. These two evaluations distinguish the benefit available from source-learned computation at insertion from improvements acquired through target-side adaptation. The pathway acts on representations produced by the receiving backbone, and target-side RL adapts its computation to this new host.

\paragraph{Sequential transfer to an adapted model.}
\label{sec:composition}
Sequential transfer inserts a source-trained pathway into a model that already contains a trained pathway. Let $\phi_1$ be the first pathway, and let $\widehat A$ and $\widehat F$ denote the frozen attention and FFN computations of the adapted model, including $\phi_1$. A second pathway $\phi_2$ acts on the residual states produced by these adapted computations:
\begin{equation}
\begin{aligned}
S &= H + \widehat A\big(N_A(H)\big), \\
U &= S + A_{\phi_2}\big(N_{A,2}(S)\big), \\
V &= U + \widehat F\big(N_F(U)\big), \\
H' &= V + F_{\phi_2}\big(N_{F,2}(V)\big).
\end{aligned}
\label{eq:sequential}
\end{equation}
The resulting order is attention--attention--FFN--FFN. The additional root-mean-square normalization (RMSNorm; \citealt{growrl_rmsnorm}) layers $N_{A,2}$ and $N_{F,2}$ are initialized by copying the corresponding normalizations and are trained with $\phi_2$. Fresh pathways start with zero output projections, while transferred pathways load learned branch components. The new normalizations contribute $2Ld$ parameters. A third round adds another attention residual step after the second attention step and another FFN residual step after the second FFN step. All earlier parameters remain frozen.

The same sequential insertion structure also supports freshly initialized pathways. Sequential transfer uses source-trained components, while fresh pathways learn new components; both use the additional normalizations described above. Section~\ref{sec:transfer_adaptation} evaluates immediate and sequential transfer without target-side optimization, as well as post-transfer adaptation. Additional implementation details are given in Appendix~\ref{app:implementation}.

% !TeX root = ../main.tex
\section{Experiments}
\label{sec:experiments}

\subsection{Experimental Setup}
\label{sec:experimental_setup}

We systematically evaluate GrowRL on language models of different sizes and types. Our experiments cover five backbones: FineMath-Llama-3B \citep{growrl_finemath}, Llama-3.2-3B \citep{growrl_llama32}, Qwen2.5-Math-1.5B and Qwen2.5-Math-7B \citep{growrl_qwenmath}, and Qwen2.5-7B \citep{growrl_qwen25}, spanning general-purpose pretrained models and mathematics-specialized models. We compare methods using shared training data, rewards, and evaluation protocols. Further experimental details are provided in Appendix~\ref{app:protocol}.

\paragraph{Reinforcement learning setup.}
We train with Dr.~GRPO \citep{growrl_drgrpo} with verifiable rewards. The training data consists of 5,585 Level 3--5 problems from the MATH training set \citep{growrl_math}. We sample eight responses per problem and reward correct final answers. The standard source-model RL budget is 500 updates. In the post-transfer adaptation experiments, we further train the transferred components for 100 target-side updates while keeping the target backbone frozen. Immediate and sequential transfer are evaluated without target-side training. Default learning rates are $10^{-6}$ for Full RL and $10^{-5}$ for parameter-efficient methods. Complete settings are provided in Appendix~\ref{app:protocol}.

\paragraph{Comparison methods.}
We compare GrowRL with full-parameter RL (Full RL), LoRA \citep{growrl_lora}, DoRA \citep{growrl_dora}, and Parallel Adapter (PA) \citep{growrl_pa}. Full RL updates all backbone parameters. LoRA and DoRA adapt the backbone projections, while PA adds a bottleneck branch alongside the FFN. GrowRL trains its added attention and FFN pathway. In immediate-transfer comparisons, Base denotes the unmodified target model and bold scores mark the best result across Base, LoRA, and GrowRL. The parameterizations are detailed in Appendix~\ref{app:protocol}.

\paragraph{Evaluation datasets.}
We evaluate models in terms of mathematical reasoning and out-of-distribution (OOD) generalization. The mathematical reasoning evaluation includes five datasets: MATH500 \citep{growrl_math500}, AMC2023 \citep{growrl_amc23}, AIME2024 \citep{growrl_aime24}, Minerva \citep{growrl_minerva}, and OlympiadBench \citep{growrl_olympiadbench}, which assess reasoning across mathematical problems of varying difficulty and form. We further use six out-of-distribution datasets to evaluate generalization after mathematical reinforcement learning: GSM8K \citep{growrl_gsm8k}, HumanEval \citep{growrl_humaneval}, MMLU-STEM and MMLU-Humanities \citep{growrl_mmlu}, ARC-Challenge \citep{growrl_arc}, and HellaSwag \citep{growrl_hellaswag}. These datasets cover mathematical and applied reasoning close to the training distribution, as well as more distant tasks involving code, knowledge, and general reasoning, allowing us to examine whether the capabilities acquired through mathematical reinforcement learning transfer beyond the training distribution.

\paragraph{Evaluation metrics.}
We consistently report sampled Pass@1 and Pass@8 \citep{growrl_humaneval} across all datasets. For each problem, we independently sample eight responses from the same generation distribution. Pass@1 reflects the average probability of success from a single sample, while Pass@8 measures the ability to obtain at least one correct answer among eight samples. Within each backbone comparison, all methods use the same prompts, sampling configurations, and answer-verification protocols. P1 and P8 denote Pass@1 and Pass@8. In the source-model RL tables, bold and underlined scores mark the best and second-best parameter-efficient results, respectively, with Full RL shown as a reference. Missing evaluations are left blank. The transfer-strength analysis uses four responses per problem on VAL300, as detailed in Appendix~\ref{app:transfer_strength}.

\subsection{Source-Model Reinforcement Learning}
\label{sec:main_results}

\paragraph{GrowRL remains consistently competitive across a broad range of models and tasks.}
Tables~\ref{tab:main_math} and~\ref{tab:main_ood} summarize source-model RL results for five backbones across 11 datasets. GrowRL achieves competitive performance in both mathematical reasoning and out-of-distribution evaluation, maintaining strong results under both Pass@1 and Pass@8. The best parameter-efficient method varies across backbones and metrics. On Qwen2.5-Math-7B, GrowRL exceeds LoRA in Pass@1 on all five mathematical benchmarks, while PA leads the parameter-efficient methods on these benchmarks under this metric. Overall, these results suggest that structural growth is effective beyond a small number of specific settings and can serve as a broadly applicable approach to reinforcement learning adaptation.

% Llama-3.2-3B DoRA source: notes/llama_dora_step150_scores_20260926.json (sampled P1/P8, update 150).
\begin{table}[!t]
\centering
\caption{\textbf{Mathematical reasoning after source-model RL.} Sampled Pass@1/Pass@8 (\%).}
\label{tab:main_math}
\begingroup
\fontsize{7.5}{8.4}\selectfont
\setlength{\tabcolsep}{2pt}
\setlength{\aboverulesep}{1pt}
\setlength{\belowrulesep}{1pt}
\renewcommand{\arraystretch}{1.00}
\begin{tabular*}{\linewidth}{@{\extracolsep{\fill}}ll*{10}{r}@{}}
\toprule
 & & \multicolumn{2}{c}{\textbf{MATH500}} & \multicolumn{2}{c}{\textbf{AMC2023}} & \multicolumn{2}{c}{\textbf{AIME2024}} & \multicolumn{2}{c}{\textbf{Minerva}} & \multicolumn{2}{c}{\textbf{Olympiad}} \\
\cmidrule(lr){3-4}\cmidrule(lr){5-6}\cmidrule(lr){7-8}\cmidrule(lr){9-10}\cmidrule(lr){11-12}
\textbf{Backbone} & \textbf{Method} & P1 & P8 & P1 & P8 & P1 & P8 & P1 & P8 & P1 & P8 \\
\midrule
FineMath-Llama-3B & Full RL & 34.65 & 65.00 & 14.06 & 40.00 & 0.42 & 3.33 & 14.02 & 33.09 & 9.51 & 28.04 \\
\cmidrule(l){2-12}
 & LoRA & 17.65 & 51.00 & 5.94 & \underline{25.00} & 0.00 & \underline{0.00} & 6.57 & 26.84 & 3.19 & 14.84 \\
 & DoRA & 7.92 & 35.20 & 2.50 & 20.00 & 0.00 & \underline{0.00} & 3.63 & 18.75 & 1.85 & 11.42 \\
 & PA & \underline{30.58} & \underline{60.60} & \underline{11.25} & \textbf{40.00} & \underline{0.42} & \textbf{3.33} & \underline{11.86} & \textbf{35.29} & \underline{7.18} & \underline{23.89} \\
 & GrowRL & \textbf{33.30} & \textbf{63.00} & \textbf{12.50} & \textbf{40.00} & \textbf{1.67} & \textbf{3.33} & \textbf{12.59} & \underline{32.72} & \textbf{8.92} & \textbf{26.26} \\
\midrule
Llama-3.2-3B & Full RL & 10.35 & 19.60 & 5.63 & 17.50 & 0.00 & 0.00 & 9.05 & 19.85 & 3.39 & 6.23 \\
\cmidrule(l){2-12}
 & LoRA & 5.53 & \underline{25.20} & 3.75 & 22.50 & \underline{0.00} & \underline{0.00} & 4.09 & 17.65 & 1.97 & \underline{10.83} \\
 & DoRA & 3.60 & 19.60 & 5.00 & \textbf{35.00} & \underline{0.00} & \underline{0.00} & 2.07 & 10.29 & 1.13 & 7.12 \\
 & PA & \underline{9.25} & \textbf{34.20} & \textbf{6.88} & \underline{25.00} & \textbf{0.83} & \textbf{6.67} & \textbf{8.23} & \textbf{23.16} & \textbf{3.65} & \textbf{13.50} \\
 & GrowRL & \textbf{9.73} & 24.20 & \underline{5.94} & \underline{25.00} & \underline{0.00} & \underline{0.00} & \underline{8.04} & \underline{21.32} & \underline{3.39} & 10.39 \\
\midrule
Qwen2.5-Math-1.5B & Full RL & 65.05 & 85.20 & 42.81 & 77.50 & 14.17 & 30.00 & 21.60 & 36.76 & 26.93 & 44.66 \\
\cmidrule(l){2-12}
 & LoRA & \underline{60.75} & \textbf{85.60} & \underline{43.44} & \underline{72.50} & \textbf{9.58} & \underline{26.67} & \underline{18.52} & 35.66 & \underline{25.24} & \textbf{45.99} \\
 & DoRA & 57.33 & 82.60 & 41.88 & \underline{72.50} & 8.33 & \textbf{30.00} & 17.51 & 35.29 & 23.33 & 43.47 \\
 & PA & 46.80 & 82.20 & 32.81 & \textbf{75.00} & 5.00 & 23.33 & 13.10 & \textbf{37.87} & 18.99 & 42.43 \\
 & GrowRL & \textbf{63.20} & \underline{85.00} & \textbf{45.94} & \textbf{75.00} & \underline{9.17} & \underline{26.67} & \textbf{19.44} & \underline{36.76} & \textbf{26.67} & \underline{44.51} \\
\midrule
Qwen2.5-Math-7B & Full RL & 67.00 & 88.40 & 58.44 & 75.00 & 19.58 & 50.00 & 23.30 & 47.06 & 29.51 & 48.52 \\
\cmidrule(l){2-12}
 & LoRA & 58.00 & 85.60 & 47.50 & \underline{82.50} & 15.83 & \underline{46.67} & 18.43 & \underline{44.49} & 24.93 & 46.88 \\
 & DoRA & 55.70 & \underline{87.20} & 47.19 & \textbf{85.00} & 13.75 & 40.00 & 16.18 & 41.18 & 23.22 & 46.29 \\
 & PA & \textbf{68.62} & \textbf{88.80} & \textbf{59.69} & \underline{82.50} & \textbf{21.25} & \textbf{53.33} & \textbf{24.36} & \textbf{45.22} & \textbf{30.86} & \underline{47.92} \\
 & GrowRL & \underline{63.75} & \underline{87.20} & \underline{54.06} & \underline{82.50} & \underline{18.75} & 43.33 & \underline{19.49} & 43.75 & \underline{27.89} & \textbf{48.66} \\
\midrule
Qwen2.5-7B & Full RL & 74.98 & 88.80 & 57.50 & 85.00 & 10.42 & 26.67 & 29.37 & 43.38 & 35.57 & 52.67 \\
\cmidrule(l){2-12}
 & LoRA & 65.68 & \underline{86.40} & 47.50 & \underline{72.50} & 12.50 & \textbf{30.00} & 25.37 & \textbf{46.69} & 28.26 & 46.88 \\
 & DoRA & 57.80 & 84.40 & 31.87 & \underline{72.50} & 6.67 & 20.00 & 18.84 & 42.65 & 23.59 & 44.96 \\
 & PA & \underline{70.10} & 85.80 & \underline{51.56} & \textbf{80.00} & \underline{12.92} & \underline{26.67} & \underline{26.61} & 43.75 & \underline{32.64} & \textbf{49.26} \\
 & GrowRL & \textbf{72.95} & \textbf{87.60} & \textbf{55.00} & \underline{72.50} & \textbf{15.83} & \textbf{30.00} & \textbf{28.03} & \underline{44.85} & \textbf{32.94} & \underline{48.52} \\
\bottomrule
\end{tabular*}
\endgroup
\end{table}

\paragraph{Structural growth approaches full-parameter reinforcement learning performance with a small number of additional parameters.}
Full RL updates all backbone parameters, whereas GrowRL optimizes only the added computational pathways. Despite a substantially smaller trainable parameter space, GrowRL achieves performance close to Full RL in many settings and matches or surpasses Full RL on some tasks and metrics. This suggests that the effective plasticity required for reinforcement learning need not be distributed throughout the entire pretrained model: compact, added computational pathways can also support substantial policy improvements.

\paragraph{Parameter-efficient reinforcement learning depends not only on how many parameters are trained, but also on how they are organized.}
LoRA, DoRA, PA, and GrowRL all restrict reinforcement learning to a small trainable parameter space, yet their performance differs substantially across backbones and tasks. GrowRL organizes its trainable parameters into complete pathways with independent attention and feed-forward computation, with competitive performance across multiple settings. These results suggest that the structural organization of trainable capacity is itself a key design dimension in parameter-efficient reinforcement learning, beyond the number of trainable parameters alone.

% Llama-3.2-3B DoRA source: notes/llama_dora_step150_scores_20260926.json (sampled P1/P8, update 150).
\begin{table}[!t]
\centering
\caption{\textbf{OOD performance after source-model RL.} Sampled Pass@1/Pass@8 (\%).}
\label{tab:main_ood}
\begingroup
\fontsize{7.5}{8.4}\selectfont
\setlength{\tabcolsep}{3pt}
\setlength{\aboverulesep}{1pt}
\setlength{\belowrulesep}{1pt}
\renewcommand{\arraystretch}{1.00}
\begin{tabular}{@{}llrr@{\hspace{9pt}}rr@{\hspace{9pt}}rr@{\hspace{9pt}}rr@{\hspace{9pt}}rr@{\hspace{9pt}}rr@{}}
\toprule
 & & \multicolumn{2}{c}{\textbf{GSM8K}} & \multicolumn{2}{c}{\textbf{HumanEval}} & \multicolumn{2}{c}{\textbf{MMLU-STEM}} & \multicolumn{2}{c}{\textbf{ARC-C}} & \multicolumn{2}{c}{\textbf{MMLU-H}} & \multicolumn{2}{c}{\textbf{HellaSwag}} \\
\cmidrule(lr){3-4}\cmidrule(lr){5-6}\cmidrule(lr){7-8}\cmidrule(lr){9-10}\cmidrule(lr){11-12}\cmidrule(lr){13-14}
\textbf{Backbone} & \textbf{Method} & P1 & P8 & P1 & P8 & P1 & P8 & P1 & P8 & P1 & P8 & P1 & P8 \\
\midrule
FineMath-Llama-3B & Full RL & 59.44 & 87.34 & 0.00 & 0.00 & 12.80 & 62.69 & 15.66 & 72.10 & 12.81 & 63.38 & 12.17 & 56.87 \\
\cmidrule(l){2-14}
 & LoRA & 31.25 & 75.44 & \underline{0.00} & \underline{0.00} & 11.56 & \underline{61.03} & 14.05 & 68.00 & 11.96 & 60.60 & \underline{12.15} & \textbf{57.07} \\
 & DoRA & 11.43 & 49.96 & \textbf{0.08} & \textbf{0.61} & 11.03 & 58.48 & 12.81 & 63.65 & 10.94 & 56.47 & 11.60 & 54.81 \\
 & PA & \underline{54.43} & \underline{86.88} & \underline{0.00} & \underline{0.00} & \underline{12.27} & 60.80 & \underline{15.18} & \underline{68.77} & \underline{12.89} & \underline{63.34} & \textbf{12.26} & 56.70 \\
 & GrowRL & \textbf{59.51} & \textbf{87.34} & \underline{0.00} & \underline{0.00} & \textbf{15.12} & \textbf{66.20} & \textbf{19.64} & \textbf{75.26} & \textbf{15.06} & \textbf{67.74} & \textbf{12.26} & \underline{57.01} \\
\midrule
Llama-3.2-3B & Full RL & 15.57 & 38.21 & 19.97 & 44.51 & 7.14 & 40.06 & 8.75 & 46.25 & 11.71 & 55.01 & 7.79 & 42.79 \\
\cmidrule(l){2-14}
 & LoRA & 8.48 & 38.59 & 13.64 & 37.20 & \underline{8.81} & \textbf{49.90} & \underline{10.65} & \underline{54.86} & \underline{11.21} & \underline{57.07} & \underline{8.91} & \textbf{48.13} \\
 & DoRA & 4.30 & 26.69 & 11.59 & 34.76 & 8.19 & 47.68 & 9.82 & 52.82 & 10.71 & 55.83 & 8.62 & \underline{46.73} \\
 & PA & \underline{13.62} & \underline{47.84} & \underline{18.60} & \textbf{42.68} & 2.92 & 18.95 & 4.58 & 27.90 & 5.41 & 32.75 & 5.78 & 32.83 \\
 & GrowRL & \textbf{16.62} & \textbf{48.22} & \textbf{20.20} & \underline{40.85} & \textbf{10.35} & \underline{49.87} & \textbf{15.31} & \textbf{62.54} & \textbf{13.61} & \textbf{59.26} & \textbf{9.46} & 44.16 \\
\midrule
Qwen2.5-Math-1.5B & Full RL & 77.23 & 94.47 & 24.31 & 68.29 & 44.57 & 83.80 & 53.30 & 93.43 & 31.89 & 84.65 & 11.84 & 45.54 \\
\cmidrule(l){2-14}
 & LoRA & \underline{72.65} & 94.31 & \underline{20.58} & \underline{65.85} & \textbf{43.85} & \underline{85.02} & \underline{49.25} & \textbf{91.89} & \underline{30.29} & \underline{84.21} & \textbf{14.02} & \underline{53.27} \\
 & DoRA & 68.85 & \underline{94.47} & 6.10 & 32.32 & 41.96 & 84.86 & 45.49 & \underline{91.64} & 28.55 & \textbf{84.72} & \underline{13.71} & \textbf{54.02} \\
 & PA & 55.93 & 93.63 & 5.79 & 34.76 & 32.67 & \textbf{86.78} & 32.35 & 89.33 & 21.49 & 80.70 & 11.84 & 51.76 \\
 & GrowRL & \textbf{74.72} & \textbf{95.22} & \textbf{31.10} & \textbf{70.73} & \underline{43.18} & 81.58 & \textbf{51.75} & \textbf{91.89} & \textbf{31.25} & 83.66 & 11.90 & 46.84 \\
\midrule
Qwen2.5-Math-7B & Full RL & 81.43 & 97.80 & 35.82 & 76.22 & 33.55 & 88.97 & 38.32 & 94.28 & 25.57 & 83.36 & 14.44 & 58.43 \\
\cmidrule(l){2-14}
 & LoRA & 70.50 & \underline{97.35} & \underline{29.95} & \textbf{75.61} & \underline{37.88} & \textbf{90.92} & \underline{43.76} & \underline{94.88} & \underline{26.53} & \underline{84.87} & \underline{16.53} & \underline{66.17} \\
 & DoRA & 67.39 & 96.82 & 28.05 & 72.56 & 33.03 & 89.33 & 41.80 & \underline{94.88} & 25.23 & 84.27 & 15.49 & 64.50 \\
 & PA & \textbf{80.35} & \textbf{97.50} & 29.12 & 73.17 & \textbf{55.48} & \underline{89.76} & \textbf{64.74} & \textbf{95.90} & \textbf{38.96} & \textbf{86.95} & \textbf{18.80} & \textbf{68.28} \\
 & GrowRL & \underline{76.30} & \textbf{97.50} & \textbf{30.95} & \underline{73.78} & 34.05 & 89.43 & 40.38 & 93.94 & 25.62 & 84.25 & 16.47 & 65.24 \\
\midrule
Qwen2.5-7B & Full RL & 91.60 & 96.89 & 69.74 & 91.46 & 68.07 & 86.61 & 87.62 & 96.42 & 58.59 & 84.95 & 40.56 & 85.71 \\
\cmidrule(l){2-14}
 & LoRA & 85.08 & 97.12 & \textbf{54.50} & \textbf{90.85} & \underline{66.09} & \underline{92.78} & 79.76 & \underline{96.84} & 51.27 & 86.82 & \underline{28.19} & \textbf{81.15} \\
 & DoRA & 78.71 & \textbf{97.50} & 26.37 & 79.88 & 56.57 & \textbf{93.64} & 69.39 & \textbf{97.35} & 43.71 & \textbf{87.23} & 26.59 & \underline{81.11} \\
 & PA & \textbf{90.51} & 96.89 & 15.17 & 60.98 & \textbf{68.06} & 90.39 & \textbf{81.76} & 95.90 & \textbf{54.93} & 82.95 & \textbf{29.26} & 79.13 \\
 & GrowRL & \underline{90.38} & \underline{97.27} & \underline{53.43} & \underline{86.59} & 63.56 & 89.46 & \underline{80.32} & \underline{96.84} & \underline{51.89} & \underline{86.91} & 21.73 & 70.96 \\
\bottomrule
\end{tabular}
\endgroup
\end{table}

\paragraph{GrowRL maintains strong reasoning capabilities beyond the training distribution.}
Although all models undergo reinforcement learning primarily on mathematical reasoning data, GrowRL maintains strong performance on out-of-distribution tasks including GSM8K, HumanEval, MMLU-STEM, ARC-Challenge, MMLU-Humanities, and HellaSwag, matching or surpassing Full RL in multiple settings. This suggests that the computational pathways learned by GrowRL are not limited to the mathematical training distribution, but can support reinforcement learning updates that are also valuable for code, knowledge, scientific reasoning, and general reasoning.

\paragraph{Structural growth remains effective across model types and scales.}
The five backbones cover both general-purpose language models and mathematics-specialized models, spanning scales of 1.5B, 3B, and 7B parameters. Despite substantial differences in their pretraining objectives, initial capabilities, and domain biases, the same GrowRL structure yields effective reinforcement learning results on all models. This indicates that the effectiveness of structural growth does not depend on a particular backbone and extends across model types and scales.

\FloatBarrier
\subsection{Transfer and Rapid Adaptation}
\label{sec:transfer_adaptation}

We evaluate immediate transfer to structurally compatible targets, post-transfer adaptation, and sequential transfer of source-trained pathways. The target backbone remains frozen throughout; immediate and sequential transfer use no target-side training.

\paragraph{Transferring computational pathways yields substantial immediate gains that extend beyond mathematical tasks.}
Table~\ref{tab:immediate_transfer} shows that GrowRL improves MATH500 performance in all three transfer directions without target-side training, outperforming direct LoRA transfer. Gains also extend to GSM8K, HumanEval, MMLU-STEM, ARC-Challenge, and MMLU-Humanities. Thus, pathways learned through mathematical RL retain value across receiving models and beyond the source training distribution.

\begin{table}[!t]
\centering
\caption{\textbf{Immediate transfer performance.} Sampled Pass@1 (\%) without target-side training.}
\label{tab:immediate_transfer}
\begingroup
\footnotesize
\setlength{\tabcolsep}{2pt}
\renewcommand{\arraystretch}{1.0}
\begin{tabular*}{\linewidth}{@{\extracolsep{\fill}}l*{10}{r}@{}}
\toprule
Method & MATH & AMC & AIME & Minerva & Olymp. & GSM8K & HEval & STEM & ARC-C & MMLU-H \\
\midrule
\multicolumn{11}{@{}l}{\textit{Qwen2.5-Math-1.5B $\rightarrow$ Qwen2.5-1.5B}} \\
Base & 4.63 & 2.50 & 0.00 & 1.10 & 1.13 & 6.12 & 3.28 & 6.58 & 7.60 & 6.25 \\
LoRA & 6.90 & 3.44 & 0.00 & 2.25 & 1.69 & 9.04 & 6.33 & 7.30 & 8.44 & 6.71 \\
GrowRL & \textbf{22.63} & \textbf{10.94} & \textbf{1.67} & \textbf{4.78} & \textbf{7.07} & \textbf{26.08} & \textbf{12.27} & \textbf{9.63} & \textbf{10.61} & \textbf{7.31} \\
\addlinespace[2pt]
\multicolumn{11}{@{}l}{\textit{FineMath-Llama-3B $\rightarrow$ Llama-3.2-3B}} \\
Base & 3.88 & 2.19 & \textbf{0.00} & 2.07 & 1.11 & 4.04 & 10.82 & 8.34 & 10.14 & 10.94 \\
LoRA & 4.45 & \textbf{3.44} & \textbf{0.00} & 3.63 & 1.39 & 7.15 & 14.63 & 9.35 & 10.63 & 11.85 \\
GrowRL & \textbf{9.63} & 2.81 & \textbf{0.00} & \textbf{5.84} & \textbf{2.26} & \textbf{15.57} & \textbf{22.26} & \textbf{9.72} & \textbf{16.21} & \textbf{15.16} \\
\addlinespace[2pt]
\multicolumn{11}{@{}l}{\textit{Qwen2.5-Math-7B $\rightarrow$ Qwen2.5-7B}} \\
Base & 32.08 & 20.31 & \textbf{3.75} & 8.87 & 10.89 & 48.54 & 24.92 & 36.44 & 46.76 & 28.66 \\
LoRA & 34.78 & 20.31 & 3.33 & 10.02 & 10.92 & 51.92 & 22.41 & 41.67 & 52.22 & 32.49 \\
GrowRL & \textbf{40.88} & \textbf{20.94} & \textbf{3.75} & \textbf{13.56} & \textbf{14.22} & \textbf{54.52} & \textbf{38.64} & \textbf{48.67} & \textbf{61.77} & \textbf{39.30} \\
\bottomrule
\end{tabular*}
\endgroup
\end{table}

\begin{figure}[!htb]
\centering
\includegraphics[width=0.95\linewidth]{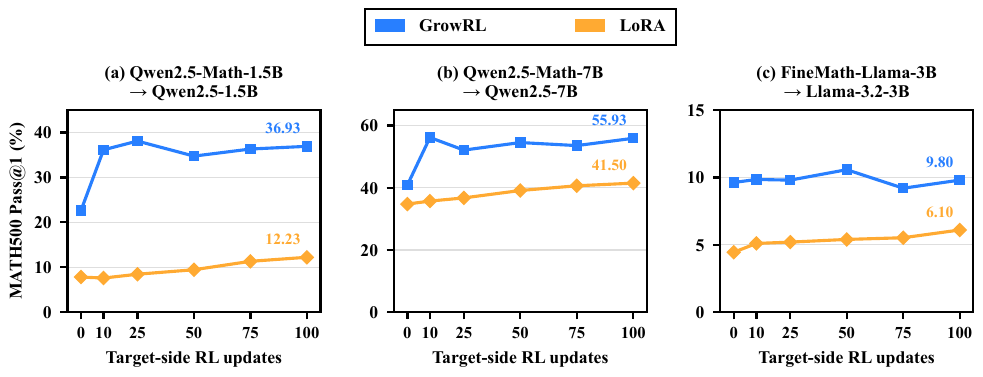}
\caption{\textbf{Rapid adaptation after transfer.} MATH500 Pass@1 with a frozen target backbone.}
\label{fig:graft_training}
\end{figure}

\paragraph{Transferred computational pathways adapt rapidly to target models with only a few reinforcement learning updates.}
In Figure~\ref{fig:graft_training}, GrowRL's MATH500 Pass@1 after 10 target-side updates is already close to its performance at update 100 on Qwen2.5-1.5B and Llama-3.2-3B, and it remains above LoRA at the measured checkpoints. Qwen2.5-1.5B improves clearly within the first 10 updates, while Llama-3.2-3B retains its immediate transfer gains. Learned pathways therefore provide an effective starting point for adaptation with a frozen target backbone.

% Place the transfer trajectory after its discussion.

\begin{samepage}
\paragraph{Sequential transfer can yield further gains without target-side training.}
Starting from each Base target, we perform up to four sequential transfers using pathway components from the same source checkpoint, without target-side optimization. Table~\ref{tab:repeated_transfer} reports the complete sweep, with 0 denoting the unmodified target. Qwen2.5-1.5B peaks at two transfers, improving Pass@1 from 23.03\% after one transfer to 42.15\%. Qwen2.5-7B improves from 39.53\% to 53.28\% over one to four transfers. Llama-3.2-3B instead peaks at one transfer and declines thereafter. Sequential transfer can thus extract additional value from source-learned computation, with the useful number of transfers depending on the receiving model.

\end{samepage}

% !TeX root = ../main.tex
% Data and construction provenance: notes/repeated_transfer_results_20260926.json
\begin{table}[!htb]
\centering
\caption{\textbf{Sequential transfer.} MATH500 Pass@1/Pass@8 (\%) without target-side training.}
\label{tab:repeated_transfer}
\begingroup
\small
\setlength{\tabcolsep}{5pt}
\renewcommand{\arraystretch}{1.0}
\begin{tabular*}{\linewidth}{@{\extracolsep{\fill}}lrrrrrr@{}}
\toprule
& \multicolumn{2}{c}{Qwen2.5-Math-1.5B} & \multicolumn{2}{c}{Qwen2.5-Math-7B} & \multicolumn{2}{c}{FineMath-Llama-3B} \\
& \multicolumn{2}{c}{$\downarrow$ Qwen2.5-1.5B} & \multicolumn{2}{c}{$\downarrow$ Qwen2.5-7B} & \multicolumn{2}{c}{$\downarrow$ Llama-3.2-3B} \\
\cmidrule(lr){2-3}\cmidrule(lr){4-5}\cmidrule(l){6-7}
Transfers & P1 & P8 & P1 & P8 & P1 & P8 \\
\midrule
0 (Base) & 4.63 & 27.80 & 31.50 & 76.60 & 4.00 & 21.20 \\
1 & 23.03 & 62.80 & 39.53 & 80.00 & \textbf{9.50} & \textbf{31.20} \\
2 & \textbf{42.15} & \textbf{75.20} & 46.08 & 81.60 & 4.93 & 15.00 \\
3 & 41.23 & 71.20 & 48.93 & 80.40 & 4.75 & 13.60 \\
4 & 33.40 & 62.80 & \textbf{53.28} & \textbf{82.20} & 4.55 & 10.80 \\
\bottomrule
\end{tabular*}
\endgroup
\end{table}

\FloatBarrier
\newpage
\subsection{Fine-Grained Analysis}
\label{sec:fine_grained_analysis}

\begin{table}[!t]
\centering
\caption{\textbf{Effect of trainable pathway components.} Sampled Pass@1 (\%) after source-model RL.}
\label{tab:pathway_trainability}
\begingroup
\footnotesize
\setlength{\tabcolsep}{2pt}
\renewcommand{\arraystretch}{1.0}

\begin{tabular*}{\linewidth}{@{\extracolsep{\fill}}llrrrrrr@{}}
\toprule
Backbone & Trainable part & Params (M) & MATH500 & AMC2023 & AIME2024 & Minerva & Olympiad \\
\midrule
Qwen2.5-Math-1.5B & Full pathway & 38.54 & \textbf{63.20} & \textbf{45.94} & \textbf{9.17} & \textbf{19.44} & \textbf{26.67} \\
 & $W_O,W_d$ only & 11.01 & 50.75 & 33.13 & 7.92 & 14.57 & 19.42 \\
\addlinespace[2pt]
Qwen2.5-Math-7B & Full pathway & 89.92 & \textbf{63.75} & \textbf{54.06} & 18.75 & 19.49 & \textbf{27.89} \\
 & $W_O,W_d$ only & 25.69 & 63.15 & \textbf{54.06} & \textbf{19.17} & \textbf{21.23} & 27.60 \\
\addlinespace[2pt]
FineMath-Llama-3B & Full pathway & 77.07 & \textbf{33.30} & \textbf{12.50} & \textbf{1.67} & \textbf{12.59} & \textbf{8.92} \\
 & $W_O,W_d$ only & 22.02 & 11.53 & 3.44 & 0.00 & 4.96 & 2.47 \\
\bottomrule
\end{tabular*}
\endgroup
\end{table}

\paragraph{Training internal transformations can improve pathway adaptation.}
Table~\ref{tab:pathway_trainability} compares \emph{Full pathway}, which trains all added projections, with \emph{$W_O,W_d$ only}, which freezes the internal projections. The latter trains only the attention output and FFN down projections that return branch features to the residual stream. Full-pathway training improves all five mathematical benchmarks on Qwen2.5-Math-1.5B and FineMath-Llama-3B. On Qwen2.5-Math-7B, it improves MATH500 and OlympiadBench, underperforms the two-projection variant on AIME2024 and Minerva, and ties on AMC2023. Learning internal transformations can therefore improve adaptation, with benefits that depend on the backbone and task.

\begin{figure}[!ht]
\centering
\includegraphics[width=0.40\linewidth,trim=0 6bp 0 8bp,clip]{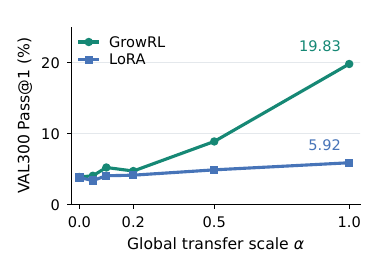}
\caption{\textbf{Effect of transfer strength.} VAL300 Pass@1 for Qwen2.5-Math-1.5B $\rightarrow$ Qwen2.5-1.5B.}
\label{fig:graft_strength}
\end{figure}

\paragraph{Increasing the contribution of learned pathways improves overall performance after transfer.}
We vary a fixed global coefficient $\alpha\in\{0,0.05,0.1,0.2,0.5,1\}$ at evaluation time, without target-side training. For GrowRL, the same $\alpha$ scales both transferred branch outputs at every layer:
\begin{equation}
A_\phi^{(\alpha)}(X)=\alpha A_\phi(X), \qquad
F_\phi^{(\alpha)}(Z)=\alpha F_\phi(Z).
\label{eq:transfer_strength}
\end{equation}
These scaled outputs replace $A_\phi(X)$ and $F_\phi(Z)$ in Equation~\eqref{eq:grow_layer}. For LoRA, each adapted target projection is $W_t^{(\alpha)}=W_t+\alpha\Delta W$, where $\Delta W$ is the transferred update including its original adapter scaling. All learned parameters remain fixed: $\alpha=0$ recovers the unmodified target model, and $\alpha=1$ applies the full transferred contribution. Figure~\ref{fig:graft_strength} shows that GrowRL's sampled Pass@1 on VAL300 generally increases with $\alpha$, with a substantially larger gain than LoRA. Local changes are not strictly monotonic, but full strength performs best among the evaluated settings, supporting the contribution of the learned pathways to transfer gains.

\FloatBarrier

% !TeX root = ../main.tex
\section{Conclusion}
\label{sec:conclusion}
GrowRL learns compact attention and feed-forward pathways on frozen backbones. Across five backbones, it achieves competitive mathematical and out-of-distribution performance, while transferred pathways provide immediate gains and rapid adaptation. These findings show that the computation acquired through an expensive RL run need not remain inseparably tied to a single model. By making this computation explicit, modular, and transferable, GrowRL offers a step toward reusable units of RL adaptation.

\bibliography{references}
\bibliographystyle{iclr2027_conference}
\clearpage
\appendix
% !TeX root = ../main.tex
\section{Experimental Details and Additional Results}
\label{app:experiments}

\subsection{Source-model training}
\label{app:protocol}
The source backbones are FineMath-Llama-3B, Llama-3.2-3B, Qwen2.5-Math-1.5B, Qwen2.5-Math-7B, and Qwen2.5-7B. Qwen2.5-1.5B \citep{growrl_qwen25} is also a transfer target. The standard source-model RL budget is 500 updates. Post-transfer adaptation trains only the transferred components for 100 target-side updates with the target backbone frozen. Immediate and sequential transfer use no target-side training.

Training uses 5,585 Level 3--5 problems from the MATH training set, with 84 prompts and eight responses per prompt per update (672 responses), temperature 1, and top-$p=1$. Binary rewards indicate final-answer correctness. We use Dr.~GRPO with group-centered rewards, no reward-standard-deviation normalization, and no KL penalty. The maximum completion length is 3,000 tokens and the model context is 4,096 tokens.

We use AdamW \citep{growrl_adamw} with $(\beta_1,\beta_2)=(0.9,0.95)$, $\epsilon=10^{-8}$, gradient clipping at 1, and no weight decay or warmup. The default learning rates are $10^{-6}$ for Full RL and $10^{-5}$ for parameter-efficient methods. This choice follows common GRPO practice: the TRL guidelines recommend $10^{-6}$ for full fine-tuning and $10^{-5}$ for LoRA \citep{growrl_trl_peft}. Forward and backward computation predominantly use bfloat16 (BF16), while master parameters and Adam optimizer states use single precision (FP32). This mixed-precision setup reduces memory and compute costs while retaining higher-precision optimizer updates; BF16 is also the standard precision in the TRL GRPO configuration \citep{growrl_trl_grpo}. We distribute optimizer states and gradients with Zero Redundancy Optimizer stage 2 (ZeRO-2) \citep{growrl_zero}.

\paragraph{Adaptation methods.}
Full RL updates all backbone parameters. LoRA applies rank-32 updates to the query, key, value, attention output, and three FFN projections, with adapter scaling $\alpha_{\mathrm{LoRA}}/r=1$ and no adapter dropout. DoRA uses the same rank and target projections and additionally trains weight magnitudes. PA uses a width-192 bottleneck with a rectified linear unit (ReLU) activation alongside the FFN; its output projection is zero-initialized. PA is a comparison method, distinct from GrowRL's sequential-transfer construction.

GrowRL adds one attention head of width 128 and one SwiGLU branch of width 128 per layer. Both output projections are zero-initialized. The backbone remains frozen. The pathway has approximately 38.5M, 77.1M, and 89.9M trainable parameters for the 1.5B, 3B, and 7B backbones, respectively.

\paragraph{Precision and the scope of the DoRA baseline.}
DoRA learns a weight magnitude as well as a low-rank direction \citep{growrl_dora}. We evaluate it under the same BF16 computation and FP32 master-parameter and optimizer-state policy as the other methods. Our checkpoint audit found that DoRA's BF16 magnitude parameters changed little or not at all despite updates to their FP32 master copies, indicating that small magnitude updates were lost during conversion to BF16. PEFT normally promotes adapter parameters to FP32 where appropriate \citep{growrl_peft_dtype}, whereas our shared configuration uses BF16 for forward computation. The reported results characterize DoRA under this common training setup, rather than its best performance with a separately optimized precision configuration.

\paragraph{Execution.}
Experiments use NVIDIA A800 GPUs with 80\,GB of memory. Training is implemented with PyTorch, Transformers, TRL, PEFT, and DeepSpeed; vLLM generates rollouts and evaluation responses. Gradient accumulation preserves the effective prompt batch when the number of training GPUs changes. Evaluation uses tensor parallelism of one.

\subsection{Evaluation protocols}
\label{app:ood}
The mathematical evaluation uses all 500 MATH500 problems \citep{growrl_math500} drawn from the MATH benchmark \citep{growrl_math}, 40 AMC2023 problems \citep{growrl_amc23}, 30 AIME2024 problems \citep{growrl_aime24}, 272 Minerva problems \citep{growrl_minerva}, and 674 problems from the English text-only mathematics subset of OlympiadBench \citep{growrl_olympiadbench}. The OOD evaluation uses 1,319 GSM8K questions \citep{growrl_gsm8k}, 164 HumanEval tasks \citep{growrl_humaneval}, 3,018 MMLU-STEM and 4,705 MMLU-Humanities questions \mbox{\citep{growrl_mmlu}}, 1,172 ARC-Challenge questions \mbox{\citep{growrl_arc}}, and 10,042 HellaSwag examples \citep{growrl_hellaswag}.

For the source-model RL and immediate-transfer comparisons, each problem receives eight sampled responses at temperature 1 and top-$p=1$. Pass@1 is the mean correctness of these responses, averaged over problems. Pass@8 is the fraction of problems with at least one correct response. It does not use majority voting. Both metrics are reported as percentages.

Qwen2.5-Math-1.5B, Qwen2.5-Math-7B, Qwen2.5-1.5B, and Qwen2.5-7B use a boxed-answer prompt for mathematical evaluation. FineMath-Llama-3B and Llama-3.2-3B use the R1-style prompt with the \texttt{</answer>} stop sequence. Methods within a backbone comparison share the prompt and decoding protocol. Mathematical answers are checked with answer verification, HumanEval programs with execution-based tests, and multiple-choice tasks with answer-letter extraction. Completion limits are 3,000 tokens for the five mathematical benchmarks, 2,048 for GSM8K, 1,024 for HumanEval, 512 for MMLU and ARC-Challenge, and 64 for HellaSwag.

Table headings use MATH for MATH500, AMC for AMC2023, AIME for AIME2024, Olymp. for OlympiadBench, HEval for HumanEval, STEM for MMLU-STEM, ARC-C for ARC-Challenge, and MMLU-H for MMLU-Humanities. These abbreviations refer to the same datasets throughout the paper.

\subsection{Transfer and post-transfer adaptation}
\label{app:transfer}
We study Qwen2.5-Math-1.5B $\rightarrow$ Qwen2.5-1.5B, Qwen2.5-Math-7B $\rightarrow$ Qwen2.5-7B, and FineMath-Llama-3B $\rightarrow$ Llama-3.2-3B. The GrowRL source pathways are taken from update 500. The LoRA adapter is held fixed across the corresponding immediate-transfer Pass@1 and Pass@8 comparisons. Immediate transfer uses zero target updates. Post-transfer adaptation runs for 100 target updates with a fresh optimizer and a frozen target backbone; only the transferred parameters are trained.

LoRA transfers the complete rank-32 adapter at its original scale. Evaluation uses the same MATH500 protocol as the source-model RL comparison. Table~\ref{tab:graft_training} lists the measured target checkpoints; line segments in Figure~\ref{fig:graft_training} connect these observations. The Qwen2.5-1.5B LoRA trajectory uses an independent eight-response evaluation of the same initial adapter weights, so its sampled update-0 score can differ from the immediate-transfer table.

\paragraph{Sequential transfer.}
Table~\ref{tab:repeated_transfer} uses the same three source--target directions, starting from each unmodified Base target. We evaluate zero through four cumulative transfers, using pathway components from the same fixed source checkpoint for each source--target pair. After the first transfer, each additional attention branch follows the preceding attention computations and each additional FFN branch follows the preceding FFN computations, as in Equation~\eqref{eq:sequential}. The additional normalizations copy the corresponding target RMSNorm weights. All parameters remain fixed, with no target-side optimization between or after transfers. Each configuration is evaluated on all 500 MATH500 problems with eight responses, temperature 1, top-$p=1$, a 3,000-token completion limit, a 4,096-token context, and the target-specific prompt described above. This sweep uses independent sampling, so its zero- and one-transfer scores can differ from Table~\ref{tab:immediate_transfer}. Bold entries mark the highest score across transfer counts for each target and metric.

\begin{table}[!t]
\centering
\caption{\textbf{Target-side adaptation after transfer.} MATH500 Pass@1 (\%) at each target update.}
\label{tab:graft_training}
\small
\begin{tabular*}{\linewidth}{@{\extracolsep{\fill}}llrrrrrr@{}}
\toprule
Target & Method & 0 & 10 & 25 & 50 & 75 & 100 \\
\midrule
Qwen2.5-1.5B & GrowRL & 22.625 & 36.150 & 38.100 & 34.750 & 36.325 & 36.925 \\
 & LoRA & 7.825 & 7.600 & 8.450 & 9.450 & 11.325 & 12.225 \\
\midrule
Qwen2.5-7B & GrowRL & 40.875 & 56.175 & 52.125 & 54.550 & 53.575 & 55.925 \\
 & LoRA & 34.775 & 35.725 & 36.750 & 39.125 & 40.650 & 41.500 \\
\midrule
Llama-3.2-3B & GrowRL & 9.625 & 9.850 & 9.800 & 10.575 & 9.200 & 9.800 \\
 & LoRA & 4.450 & 5.100 & 5.200 & 5.400 & 5.525 & 6.100 \\
\bottomrule

\end{tabular*}
\end{table}

\subsection{Additional immediate-transfer results}
\label{app:immediate_transfer_metrics}
Table~\ref{tab:immediate_transfer_pass8} complements the Pass@1 comparison in Table~\ref{tab:immediate_transfer}. Base denotes the unmodified target model. Each Pass@8 entry is computed from the same eight responses per problem as its corresponding Pass@1 entry, using the same donor and target checkpoint. Bold scores indicate the best result within each transfer direction, including ties.
\begin{table}[!t]
\centering
\caption{\textbf{Immediate transfer performance.} Sampled Pass@8 (\%) without target-side training.}
\label{tab:immediate_transfer_pass8}
\begingroup
\footnotesize
\setlength{\tabcolsep}{2pt}
\renewcommand{\arraystretch}{1.0}
\begin{tabular*}{\linewidth}{@{\extracolsep{\fill}}l*{10}{r}@{}}
\toprule
Method & MATH & AMC & AIME & Minerva & Olymp. & GSM8K & HEval & STEM & ARC-C & MMLU-H \\
\midrule
\multicolumn{11}{@{}l}{\textit{Qwen2.5-Math-1.5B $\rightarrow$ Qwen2.5-1.5B}} \\
Base & 27.80 & 20.00 & 0.00 & 5.88 & 7.72 & 36.39 & 19.51 & 40.92 & 45.22 & 39.28 \\
LoRA & 35.80 & 20.00 & 0.00 & 12.87 & 10.53 & 47.54 & 31.10 & 45.00 & 48.89 & 41.64 \\
GrowRL & \textbf{65.20} & \textbf{40.00} & \textbf{6.67} & \textbf{20.22} & \textbf{26.41} & \textbf{80.29} & \textbf{48.78} & \textbf{53.51} & \textbf{56.74} & \textbf{43.49} \\
\addlinespace[2pt]
\multicolumn{11}{@{}l}{\textit{FineMath-Llama-3B $\rightarrow$ Llama-3.2-3B}} \\
Base & 20.00 & 17.50 & \textbf{0.00} & 12.13 & 6.97 & 23.88 & 33.54 & 47.32 & 54.44 & 56.17 \\
LoRA & 23.20 & \textbf{22.50} & \textbf{0.00} & 15.44 & 8.46 & 35.86 & 42.07 & \textbf{52.05} & 55.20 & 58.89 \\
GrowRL & \textbf{30.60} & 17.50 & \textbf{0.00} & \textbf{18.75} & \textbf{9.64} & \textbf{55.34} & \textbf{48.78} & 44.63 & \textbf{63.14} & \textbf{63.93} \\
\addlinespace[2pt]
\multicolumn{11}{@{}l}{\textit{Qwen2.5-Math-7B $\rightarrow$ Qwen2.5-7B}} \\
Base & 76.80 & 57.50 & \textbf{20.00} & 34.93 & 36.80 & 95.68 & 78.05 & 89.86 & 96.84 & 84.57 \\
LoRA & 76.60 & \textbf{62.50} & 16.67 & 36.40 & 35.61 & 95.22 & 79.27 & 91.98 & 96.84 & 86.27 \\
GrowRL & \textbf{80.00} & 55.00 & 16.67 & \textbf{41.91} & \textbf{37.98} & \textbf{96.06} & \textbf{87.20} & \textbf{92.05} & \textbf{97.44} & \textbf{88.20} \\
\bottomrule
\end{tabular*}
\endgroup
\end{table}

\subsection{Trainable components and transfer strength}
\label{app:transfer_strength}
The component ablation in Table~\ref{tab:pathway_trainability} compares \emph{Full pathway}, which trains the complete added pathway, with \emph{$W_O,W_d$ only}, which trains only the added attention output projection $W_O$ and FFN down projection $W_d$. In the latter variant, $W_Q,W_K,W_V,W_g,W_u$ remain at their random initialization. The backbone, including the language-model output head, is frozen in both variants. Both variants contain an attention branch and an FFN branch; their trainable parameter counts are reported in the table. Bold scores indicate the better result within each backbone.

The transfer-strength experiment uses Qwen2.5-Math-1.5B $\rightarrow$ Qwen2.5-1.5B without target-side training. We vary the contribution of the transferred components with $\alpha\in\{0,0.05,0.1,0.2,0.5,1\}$. As defined in Equation~\eqref{eq:transfer_strength}, GrowRL scales both branch outputs at every layer; LoRA scales each transferred adapter update. All learned parameters remain fixed during the sweep.

This analysis uses VAL300, a fixed set of 300 Level 3--5 problems sampled from the MATH test split, stratified by difficulty (92 Level 3, 99 Level 4, and 109 Level 5 problems). We verified that VAL300 has no overlap with the 5,585 RL training problems by either problem ID or normalized problem text. Each problem receives four sampled responses at temperature 1. The plotted mean binary reward is sampled Pass@1 on this validation set.

\FloatBarrier
% !TeX root = ../main.tex
\section{Method and Implementation Details}
\label{app:method_details}

\subsection{Function preservation and initial gradients}
\label{app:initialization}
At initialization, $W_O=W_d=0$ makes both added residual corrections identically zero. Inducting over the attention and feed-forward sublayers in Equation~\eqref{eq:grow_layer} shows that every hidden state, and hence the policy distribution, equals that of the pretrained model in exact arithmetic. The same argument applies to inserting a fresh zero-output pathway into an adapted model. A transferred learned pathway instead uses nonzero output projections and can change the receiving model's policy immediately.

For either branch, write its output as $B_\psi(X)W_{\mathrm{out}}$ and let $D$ denote the loss gradient at that output. Then
\begin{equation}
\nabla_{W_{\mathrm{out}}}\mathcal L=B_\psi(X)^\top D,
\qquad
\nabla_\psi\mathcal L=J_\psi[\operatorname{vec}B]^\top\operatorname{vec}\!\left(DW_{\mathrm{out}}^\top\right).
\end{equation}
Here $J_\psi[\operatorname{vec}B]$ is the Jacobian of the vectorized internal features with respect to the internal projection parameters. Thus, at zero output initialization, the internal projection gradients vanish, while the output gradients can be nonzero. Output changes allow internal learning on subsequent updates. The $W_O,W_d$-only control keeps the internal projection matrices at their random initialization; its inputs and activations can still change as earlier branches learn.

\subsection{Complete RL objective}
\label{app:rl_objective}

We train the pathway using Dr.~GRPO \citep{growrl_drgrpo}. Let $\pi_{\theta,\phi}$ denote the policy with frozen backbone parameters $\theta$ and trainable pathway parameters $\phi$. For each prompt $x$, we sample $G$ completions $y_1,\ldots,y_G$ and obtain binary answer-verification rewards $R_i\in\{0,1\}$. Each token in completion $i$ receives the group-centered advantage
\begin{equation}
a_i = R_i-\frac{1}{G}\sum_{j=1}^{G}R_j,
\qquad
\rho_{i,t}(\phi) =
\frac{\pi_{\theta,\phi}(y_{i,t}\mid x,y_{i,<t})}
     {\pi_{\mathrm{old}}(y_{i,t}\mid x,y_{i,<t})},
\label{eq:advantage_ratio}
\end{equation}
where $\pi_{\mathrm{old}}$ is the actor before the update. Following clipped policy optimization \citep{growrl_ppo}, we minimize the following per-prompt loss, averaged over the prompt batch:
\begin{equation}
\mathcal{L}(\phi) = -\frac{1}{G T_{\max}}
\sum_{i=1}^{G} w_i \sum_{t=1}^{|y_i|}
\min\!\left[\rho_{i,t}(\phi)a_i,\,
\operatorname{clip}\big(\rho_{i,t}(\phi),1-\epsilon,1+\epsilon\big)a_i\right].
\label{eq:drgrpo}
\end{equation}
Here $T_{\max}$ is a fixed completion-length limit. We use $G=8$, $T_{\max}=3000$, and $\epsilon=0.2$, without reward-standard-deviation normalization or a KL penalty. The fixed denominator avoids dividing each completion's contribution by its own length.

The detached weight $w_i$ accounts for differences between the rollout engine and the training actor. If $b$ denotes the rollout policy, the sequence importance ratio is $s_i=\pi_{\mathrm{old}}(y_i\mid x)/b(y_i\mid x)$. Our GrowRL runs use $w_i=s_i\mathbf{1}[s_i\leq 3]$; identical rollout and actor probabilities give $w_i=1$. This sequence-level correction is separate from the token-level clipping in Equation~\eqref{eq:drgrpo}. The same objective trains newly inserted pathways after transfer or composition, with only the newly added pathway parameters updated.

\subsection{Implementation}
\label{app:implementation}
For source-model RL, the added attention and FFN branches can be represented by concatenating their channels and output projections with those of the backbone. Zero-output padding heads preserve the backbone's grouped-query attention layout \citep{growrl_gqa} where required, while each layer has one effective added attention head. Frozen backbone and padding slices are restored after optimizer steps. Sequential transfer retains the explicit residual steps and additional normalization layers in Equation~\eqref{eq:sequential}.

Forward computation uses BF16, with FP32 master parameters and optimizer states. The parameter counts in the paper measure trainable pathway parameters; dense storage and optimizer allocation can be larger. Frozen backbone parameters remain unchanged, while the hidden states can change as the pathway learns.

\end{document}